\documentclass[final]{cvpr}

\usepackage{times}
\usepackage{epsfig}
\usepackage{graphicx}
\usepackage{amsmath}
\usepackage{amssymb}

\usepackage{booktabs}
\usepackage{threeparttable}
\usepackage{multirow}

\usepackage{algorithm}
\usepackage{algpseudocode}

\algnewcommand{\algorithmicforeach}{\textbf{for each}}
\algdef{SE}[FOR]{ForEach}{EndForEach}[1]
{\algorithmicforeach\ #1\ \algorithmicdo}% \ForEach{#1}
{\algorithmicend\ \algorithmicforeach}% \EndForEach

\usepackage{color}
\usepackage[pagebackref=true,breaklinks=true,colorlinks,bookmarks=false]{hyperref}

\def\cvprPaperID{4039} % *** Enter the CVPR Paper ID here
\def\confYear{CVPR 2021}

\begin{document}

%%%%%%%%% TITLE
\title{Progressive Stage-wise Learning for Unsupervised \\ Feature Representation Enhancement}

\author{Zefan Li$^{14}$\thanks{Work done while visiting Johns Hopkins University.}\\
%Shanghai Jiao Tong University\\
{\tt\small leezf@sjtu.edu.cn}
\and
Chenxi Liu$^2$\thanks{Now at Waymo.}\\
%Johns Hopkins University\\
{\tt\small cxliu@waymo.com}
\and
Alan Yuille$^2$\\
%Johns Hopkins University\\
{\tt\small alan.l.yuille@gmail.com}
\and
Bingbing Ni$^{14}$\thanks{Corresponding author.}\\
%Shanghai Jiao Tong University\\
{\tt\small nibingbing@sjtu.edu.cn}
\and
Wenjun Zhang$^1$\\
%Shanghai Jiao Tong University\\
{\tt\small zhangwenjun@sjtu.edu.cn}
\and
Wen Gao$^3$\\
%Peking University\\
{\tt\small wgao@pku.edu.cn}
\and
$^1$Shanghai Jiao Tong University \quad $^2$Johns Hopkins University
\quad $^3$Peking University\\
$^4$MoE Key Lab of Artificial Intelligence, AI Institute, Shanghai Jiao Tong University\\
}

% \author{Zefan Li^1\thanks{Work done while at Johns Hopkins University.}\\
% Shanghai Jiao Tong University\\
% {\tt\small leezf@sjtu.edu.cn}
% \and
% Chenxi Liu\thanks{Now at Waymo.}\\
% Johns Hopkins University\\
% {\tt\small cxliu@waymo.com}
% \and
% Alan Yuille\\
% Johns Hopkins University\\
% {\tt\small alan.l.yuille@gmail.com}
% \and
% Bingbing Ni^1\thanks{Corresponding author.}\\
% Shanghai Jiao Tong University\\
% {\tt\small nibingbing@sjtu.edu.cn}
% \and
% Wenjun Zhang\\
% Shanghai Jiao Tong University\\
% {\tt\small zhangwenjun@sjtu.edu.cn}
% \and
% Wen Gao\\
% Peking University\\
% {\tt\small wgao@pku.edu.cn}
% \and
% $^1$MoE Key Lab of Artificial Intelligence, AI Institute, Shanghai Jiao Tong University
% }

\maketitle

%%%%%%%%% ABSTRACT
\begin{abstract}
	Unsupervised learning methods have recently shown their competitiveness against supervised training. Typically, these methods use a single objective to train the entire network. But one distinct advantage of unsupervised over supervised learning is that the former possesses more variety and freedom in designing the objective. In this work, we explore new dimensions of unsupervised learning by proposing the \textbf{P}rogressive \textbf{S}tage-wise \textbf{L}earning (\textbf{PSL}) framework.
	For a given unsupervised task, we design multi-level tasks and define different learning stages for the deep network.
	Early learning stages are forced to focus on low-level tasks while late stages are guided to extract deeper information through harder tasks. We discover that by progressive stage-wise learning, unsupervised feature representation can be effectively enhanced. Our extensive experiments show that PSL consistently improves results for the leading unsupervised learning methods.
	%This stage-wise progressive learning closes the performance gap between supervised and unsupervised learning algorithms.
	
\end{abstract}

\section{Introduction}
\thispagestyle{empty}
%Supervised learning is reaching its saturation due to its requirement for intensive manual annotation of millions of data samples.  
Aiming at learning features from label-free data, unsupervised representation learning, including self-supervised learning, is an important problem to study. 
Many efforts have been made, to bridge the performance gap between supervised and unsupervised learning algorithms. These methods can be roughly divided into two categories: i) handcrafted pretext tasks, that learns data-level invariant features (e.g., jigsaw puzzle~\cite{Jigsaw}, image rotation~\cite{Rotation}, image colorization~\cite{Colorization2015}) 
and ii) contrastive visual representation learning, which learns the similarity and dissimilarity between data pairs~\cite{SwAV, SimCLR, MocoV2, Moco}. For approaches using pretext tasks, they usually generate pseudo labels based on some data attributes and learn visual features through corresponding objective functions of the pretext tasks. Therefore, the final performance of these approaches is highly related to how the pretext tasks were initially designed. Most pretext tasks are designed heuristically, limiting the quality of learned representation. 
Contrastive learning methods usually generate positive/negative sample pairs through a set of image transformations and learn visual representation by bringing positive sample pairs closer while pushing  negative sample pairs away from each other. 
	\begin{figure}[t]
		\begin{center}
			\includegraphics[width=1\linewidth]{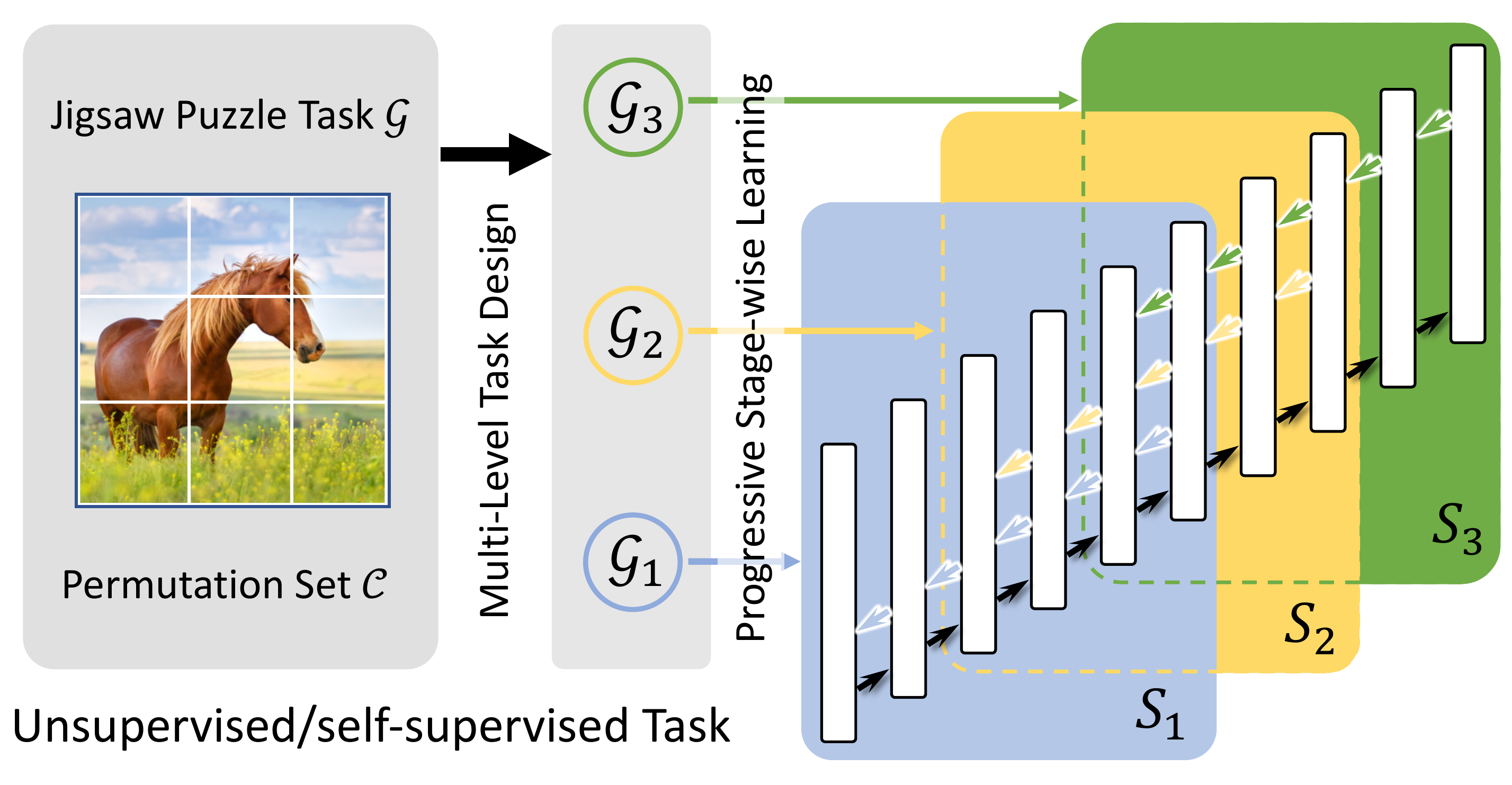}
		\end{center}
		\caption{We present the framework of the proposed Progressive Stage-wise Learning (\textbf{PSL}) algorithm, aiming for improving unsupervised/self-supervised task. We take the jigsaw puzzle task $\mathcal{G}$ for example. We first do multi-level task partition $\mathcal{G}\rightarrow\{\mathcal{G}_1, \mathcal{G}_2, \mathcal{G}_3\}$ with an increased task complexity and perform progressive stage-wise training for different learning stages of the network. The black arrow denotes forward pass while colored arrow represents the backward pass of each learning stage (i.e., $S_1$, $S_2$, and $S_3$). }
		\label{fig:1}
	\end{figure}
The design of the contrastive loss and the configuration of image transformations are essential to the quality of the resulting networks. Those methods have shown great promise in the area of unsupervised learning, achieving state-of-the-art results~\cite{Contrastive2006, Contrastive2014, CPC, AMDIM}.
Some recent methods show it is possible for unsupervised learned features to surpass supervised learning in some downstream applications~\cite{SwAV}. 
However, performance gaps still exists between unsupervised and supervised learning methods in most cases~\cite{Scaling}. 
%Though great progress has been made in the area of unsupervised learning, there is still a performance gap, compared with supervised learning. 
Therefore, how to fully explore the potential of unsupervised learning and improve the learning quality is a valuable topic.

%In this work, we aim for further improving the quality of unsupervised learning, or doing unsupervised learning more effectively. 
Instead of designing a new pretext task or a better contrastive learning loss, we try to look into this problem from a new perspective. 
As curriculum learning~\cite{Curriculum} suggests, when dealing with a complex learning target, learning things progressively can be very useful. 
Indeed, as humans, we learn visual concepts from easy to hard, and from elementary to fine-grained.
Similarly, learning high-quality feature representations in an unsupervised manner is a challenging task, and may benefit from such ideas.
In this paper, we propose \textit{\textbf{PSL}}, a progressive stage-wise learning framework for unsupervised visual representation learning. 

As presented in Fig~\ref{fig:1}, for a given unsupervised learning task $\mathcal{G}$ (e.g., the jigsaw puzzle pretext task), we first do \textit{\textbf{multi-level task design}} $\mathcal{G}\rightarrow\{\mathcal{G}_1, \mathcal{G}_2, \mathcal{G}_3\}$ with an increased task complexity. 
Then, a \textit{\textbf{stage-wise network partition}} is performed to get early/mid/late stages (i.e., $S_1$, $S_2$ and $S_3$).
 Each learning stage is assigned with a task, following the principle of easy-to-hard. Then, a \textit{\textbf{stage-wise training}} is performed. The training of lower stages become much easier as they focus on more simple tasks. The feature representations learned in upper stages are of better quality because they are trained upon the learning experience of former tasks. 
Our starting point is to design PSL as a plug-in learning method, which can be applied in any unsupervised learning scenarios under proper multi-level task design. 
We validate the effectiveness of the proposed PSL framework by evaluating our method on several unsupervised/self-supervised tasks (e.g., the jiasaw puzzle~\cite{Jigsaw} and image rotation~\cite{Rotation} pretext task and contrastive learning~\cite{SimCLR}) and present results on linear classification, semi-supervised learning and transfer learning. 

In general, the contributions of this paper can be summarized as follows:
\begin{itemize}
	\item PSL creates new dimensions for unsupervised learning research. Specifically, this includes task series, network partitions, and stage-wise training.
	\item PSL is design to be a general framework, that can be applied to multiple unsupervised learning tasks belonging to either pretext tasks or contrastive learning (e.g., jigsaw puzzle, image rotation, and SimCLR). 
	\item By experiments of downstream applications (e.g., semi-supervised learning, transfer learning), we show that the feature representations learned by PSL consistently achieve better quality than the original unsupervised task.
\end{itemize}

\section{Related Work}
%The recent success of neural networks boosts massive research in learning feature representations without  manually annotated label supervision. Our work is mainly related to the following research areas.
Our method falls in the area of unsupervised visual representation learning. We first revisit two categories of unsupervised learning method. Then, we review methods involving self-paced learning and local network training, which give inspiration in our PSL training scheme. 

 \vspace{-0.3cm}
 \paragraph{Handcrafted Pretext Tasks.}
 Many self-supervised methods use a handcrafted pretext task to learning visual representations. Typically, a pretext task involves predicting an explicit property of an image transformation and the network is then trained to learn the feature representation. The quality of the learned representation is highly related to these tasks (e.g., predicting context~\cite{ContextPrediction}, image rotation~\cite{Rotation}, image colorization~\cite{Colorization2015, Colorization2016A, Colorization2016B, Colorization2016C, Colorization2017}, jigsaw puzzle~\cite{Jigsaw, Jigsaw2019, Scaling} and visual counting~\cite{Count}). Instead of designing a new pretext task, we propose a plug-in method to enhance the self-supervised learning, which can be used collaboratively with many pretext tasks.
 %Extensive experiments show that, our PSL training framework improves the quality of the learned feature representations for a given pretext task. 
 
 \vspace{-0.3cm}
 \paragraph{Contrastive learning.}
 Unsupervised contrastive learning recently attracts lots of attention, for achieving state-of-the-art results on ImageNet~\cite{SimCLR, Moco, SwAV}. Typically, a contrastive learning method learns feature representations by contrasting positive pairs against negative pairs, which is firstly proposed by Hadsell \textit{et al.}~\cite{Contrastive2006}. Then, Dosovitskiy \textit{et al.}~\cite{Contrastive2014} propose to represent each instance with a parametric vector. Later, the concept of memory bank, which stores the information of instance class representation, is adopted and developed further in many works~\cite{LA2019, tian2019contrastive, misra2020self}.  
 %However, maintaining a memory bank during training can be computationally expensive, which is a potential drawback. To address this issue, the memory bank is then replaced by Momentum Encoder, which generates a dictionary as a queue of encoded keys~\cite{Moco}. 
 %All methods above use instance-level contrastive algorithms, which use a similarity metric to learn the similarity between samples. 
 Besides, there are many clustering-based methods~\cite{caron2018deep, caron2019unsupervised, asano2019self,  gidaris2020learning, huang2019unsupervised, xie2016unsupervised}. 
 %For example, Caron \textit{et al.}~\cite{caron2018deep, caron2019unsupervised} use k-means assignments as the pseudo-labels for supervision and Asano \textit{et al.}~\cite{asano2019self} formulate this problem as an instance of the optimal transport problem. The performance of contrastive learning is influenced by multiple factors, e.g., how data pair are sampled~\cite{SimCLR}, how cluster information are learned~\cite{SwAV} and how the contrastive loss are formulated~\cite{tschannen2019mutual}.
 Our PSL training scheme can also fit into contrastive learning methods, bringing improvement to the quality of the learned feature representation. 
 \paragraph{Self-paced Learning}
 Many self-paced learning methods simulate the learning process of ``easy-to-hard''~\cite{selfpaced1,selfpaced2,selfpaced3, selfpaced4}.  \cite{selfpaced1} incorporates self-paced learning into deep clustering methods by controlling the number of selected data samples. \cite{selfpaced2} uses a self-paced learning strategy by identifying reliable and unreliable clusters to improve the accuracy in the re-clustering step. These methods are based on data-level information by defining easy/hard data samples while our method focus on the task-level design.
 \vspace{-0.3cm}
 \paragraph{Local Network Learning.} 
 The end-to-end training protocol inaugurated a new era in deep learning. Some works jump out of traditional forward-backward training mode, with inspiration from neuroscience. An early research~\cite{linsker1988self} shows that the way of the brain processing its perceptions is to maximally preserve the information contained in each layer. Several methods try to explore greedy layer-wise training schemes~\cite{bengio2006greedy, belilovsky2019greedy}, in which the possibility of scaling this scheme to ImageNet is discussed. Later, GIM~\cite{GIM} argues that with greedy self-supervised training, end-to-end propagation of a supervised loss is not necessary. 
 %With only local learning rules, those methods are basically inferior than those state-of-the-art networks trained in end-to-end propagation, especially on large scale datasets. 
 Based on GIM~\cite{GIM}, LoCo~\cite{Loco} improves the performance of local contrastive learning. Inspired by these local training strategy, we propose a stage-wise training algorithm and improve the performance in multiple unsupervised learning task.

\section{Method}
\subsection{Overview}
This work aims to introduce an unsupervised pre-training strategy.  Many previous methods focus on designing handcrafted pretext tasks in a self-supervised learning setting~\cite{Jigsaw,Rotation,Colorization2015}. Correspondingly, the results are highly related to how the pretext tasks are designed. Another type of work focuses on exploring the potential behind contrastive learning. These works concentrate on learning similar/dissimilar representations from organized similar/dissimilar data pairs. Both kinds are trying to uncover internal information of unlabeled data. 
%However, several problems may be encountered. Firstly, to meet the demand for complex applications, the complexity of neural networks (e.g., depth and width) may be increased. Second, the data size and diversity keep expanding (especially unlabeled data). Unsupervised training becomes more and more challenging. 
As unsupervised learning becomes more and more important and challenging, how to take the best advantage of existing unsupervised learning ways is vital. In another word, how to do unsupervised learning more effectively?

%as the complexity of both data and neural networks increases, the unsupervised task becomes much more difficult. To increase the model performance, deeper neural networks may be designed and a stronger unsupervised/self-supervised task may be applied. 

In this work, we provide a new dimension in enhancing the unsupervised learning representation. Inspired by curriculum learning,  we try to guide the neural network to learning feature representations in a progressive way (e.g., from easy tasks to hard tasks, from low-level features to high-level).  To do so, we introduce our progressive stage-wise learning (PSL) framework for unsupervised learning. 

\subsection{Progressive Stage-wise Learning}\label{sec:PSL}

\begin{figure*}[t]
	\begin{center}
		\includegraphics[width=0.9\linewidth]{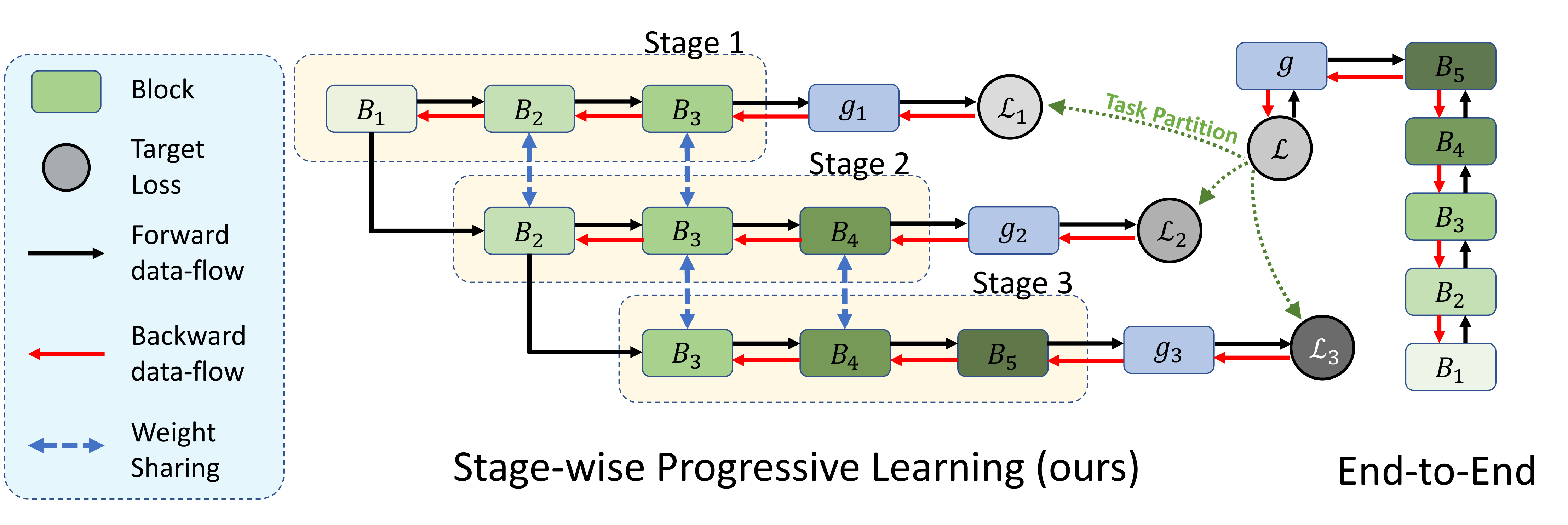}
	\end{center}
	\caption{We present the detail of the proposed Stage-wise Progressive Learning framework. In the right is the end-to-end learning scheme while we present PSL in the middle. $g$ and $\{g_i\}^3_{i=1}$ are projection heads, mapping the intermediate representation to the target feature space. After the training is completed, we throw away the projection heads and use the backbone network for downstream tasks. }
	\label{fig:framework}
\end{figure*}

In this sub-section, we explain how our progressive stage-wise learning (PSL) works. Suppose we have a learning target $\mathcal{G}$, which can be an unsupervised contrastive learning task or any pretext task in the self-supervised learning setting. We use a neural network with a block-based architecture (designed by stacking blocks vertically, such as ResNet~\cite{ResNet} and Inception~\cite{Inception}). What we do can be summarized into the following steps:

%As shown in the left part of Fig~\ref{fig:framework}, a common practice is to perform end-to-end training with respect to learning target $\mathcal{G}$. What PBL does is shown in the right part of Fig~\ref{fig:framework}. 

\paragraph{Multi-level Task Design} 
	Firstly, for the given learning target $\mathcal{G}$, we design a series of learning tasks that share a similar form (e.g. the same pretext task) but with different task complexity. We sort these tasks according to their complexity: $\{\mathcal{G}_1, \mathcal{G}_2, \mathcal{G}_3\}$, where $\mathcal{G}_3$ is more complex than $\mathcal{G}_2$, and so is $\mathcal{G}_2$ compared to $\mathcal{G}_1$ . We use $\mathcal{L}_1, \mathcal{L}_2, \mathcal{L}_3$ to represent the corresponding loss function. Instead of focusing on the hardest task (e.g., $\mathcal{G}_3$) at the very beginning, we enable easier learning targets in the early learning stage.
	By doing so, we can train the neural network in a progressive learning manner, which turns out to be more efficient in many different unsupervised settings.
	We introduce our multi-level task design for specific tasks in Sec~\ref{sec:task}. 
	
	\paragraph{Stage-wise Network Partition} 
	In this step, we define different learning stages, which are basically determined by the layer depth.	
	We take ResNet-50 for an example.  Based on the resolution of the feature map, we divide all layers into five large block: $B_1$, $B_2$, $B_3$, $B_4$ and $B_5$, where $B_1$ represents the layer \texttt{conv1},  $B_2$ consists of all layers named \texttt{conv2\_x}, and so on\footnote{Here, all layers within the same block shares the same feature map resolution. The layer name \texttt{conv1} and \texttt{conv2\_x} are inherited from the notation of ResNet~\cite{ResNet}. }.
	Further, we group every three consecutive blocks into one learning stage: Stage $S_1$ consists of $B_1$, $B_2$ and $B_3$; Stage $S_2$ consists of $B_2$, $B_3$ and $B_4$; Stage $S_3$ consists of $B_3$, $B_4$ and $B_5$. By doing so, we get three stages ($S_1$, $S_2$ and $S_3$), representing lower, mid and upper learning stages of the network. Notice that there are overlaps between each two learning stages, which is discussed more in Sec~\ref{sec:StageWiseTraining}.
	
	\paragraph{Progressive Network Learning}
	After we get a series of multi-level tasks and a proper stage partition, we begin the training process. As shown in Fig~\ref{fig:framework}, the traditional end-to-end training protocol, forces the whole neural network to learn the final target directly. Instead, we train the neural network progressively in a stage-wise manner.
	More specifically, we set a different learning target for each stage, following the principle of easy-to-hard. 
	We force lower layers of the network (e.g., layers in $S_1$) to learn low-level features by solving easier tasks (e.g., $\mathcal{G}_1$). As the lower part of the network learns a good feature representation for the easy task, we increase the task complexity for the mid and upper stages by targeting at the task $\mathcal{G}_2$ and  $\mathcal{G}_3$. By doing so, we naturally guide the network to gain better feature representation ability as the layer goes deeper. The benefits are three-fold. Firstly, the lower stage only has to focus on the easy task, which leads to easier training. Secondly, the upper stage can take advantage of the lower stages through weight sharing when dealing with a harder task. Thirdly, the backward propagation path is much shorter within each learning stage compared to end to end training. Without gradient error accumulation, the overall training can be much more efficient and effective.
	
	More details of our framework are shown in Fig~\ref{fig:framework}. 
	The training targets $\{\mathcal{G}_1, \mathcal{G}_2, \mathcal{G}_3\}$ are assigned to stages $\{S_1, S_2, S_3\}$ respectively. Notice that different from end-to-end training, 
	backward gradients only flow back to layers in the same stage.

%
%\begin{figure*}[t]
%\begin{center}
%%\fbox{\rule{0pt}{2in} \rule{0.9\linewidth}{0pt}}
%   \includegraphics[width=0.8\linewidth]{fig/1.pdf}
%\end{center}
%   \caption{TO DO}
%\end{figure*}

\subsection{Multi-level Task Design Cases}\label{sec:task}
	In this sub-section, we introduce our multi-level task design for several different unsupervised/self-supervised learning tasks.
	\subsubsection{Multi-level Jigsaw Puzzle}\label{sec:task_Jigsaw}

		The jigsaw puzzle task~\cite{Jigsaw} was first introduced to learn visual representations from unlabeled data, which turns out to be very useful in many downstream tasks, such as detection and classification. To create a jigsaw puzzle, a $225 \times 225$ pixel window is randomly cropped from an image. Then, the whole window is divided into a $3 \times 3$ grid, leading to nine $75 \times 75$ pixel cells. For each cell, a $64 \times 64$ pixel tile is picked randomly. Then an index permutation is sampled (e.g., $\{1,2,3,4,5,6,7,8,9\}\rightarrow \{3, 5, 7, 8, 4, 6, 9, 2, 1\}$) from a pre-defined permutation set $\mathcal{C}$. 
		The obtained 9 tiles are reordered according to the permutation. Finally, the reordered tiles of the puzzle are stacked along the channels and fed to the neural network to predict the permutation, which is usually a classification task.  
		There are several factors that influence the complexity of the jigsaw puzzle. 
		According to~\cite{Jigsaw}, the permutation set $\mathcal{C}$ influences the jigsaw task from two aspects: the set cardinality of $\mathcal{C}$ and the element similarity of $\mathcal{C}$. Generally, the difficulty of the jigsaw task increases as the set cardinality increases or the element similarity decreases. Under the original task setting, there are $9!=362880$ different permutations in total for every 9 tiles. Therefore, it is nearly impossible to include all permutations in the permutation set. Previous methods~\cite{Jigsaw, Jigsaw2019} usually define a permutation set with a fixed size (e.g., 1000) in advance. 
		
		 In this work, we design multi-level jigsaw puzzle tasks by changing cardinality of $\mathcal{C}$. As shown in Table~\ref{tab:jigsaw}, we generate three permutation sets $\mathcal{C}_1, \mathcal{C}_2, \mathcal{C}_3$ with cardinality 500, 1000, and 2000. Notice that $\mathcal{C}_1\subseteq \mathcal{C}_2\subseteq \mathcal{C}_3$. We keep the average hamming distance of each permutation set around 8.0 so that the element similarity within each set stays in the same level.
		 The task complexity increases as permutation gets bigger. 
		 
		 \begin{table}
		 \begin{center}
		 \begin{tabular}{c|c|c|c}
		 \hline
		 Task & Set$^\dagger$ & Cardinality & Hamming  \\
		 \hline
		 $\mathcal{G}_1$ & $\mathcal{C}_1$ & 500   & 	$\sim$8.0  \\
		 $\mathcal{G}_2$ & $\mathcal{C}_2$ & 1000   &  $\sim$8.0  \\
		 $\mathcal{G}_3$ & $\mathcal{C}_3$ & 2000 &  $\sim$8.0  \\
		 \hline
		 \end{tabular}
		 \end{center}
		 \caption{Multi-level task design for jigsaw puzzle. Set$^\dagger$ denotes the permutation set.}
		 \label{tab:jigsaw}
		 \end{table}
		 
		 In addition to the change of cardinality of permutation sets, there are other methods to control the task complexity. 
		For example, one can increase or decrease the size of the grid (e.g. from $3 \times 3$ to $2 \times 2$ and $4 \times 4$). However, the resulting difficulty gap between adjacent tasks is too larger under our multi-level jigsaw puzzle design. \cite{Scaling} reported that increasing the number of patches (i.e. from 9 to 16) does not necessarily result in a higher quality representation.
		Therefore, we do not adopt this scheme in this work.
		
		\subsubsection{Multi-level Image Rotation}\label{sec:task_Rotation}
		The Image rotation task was firstly designed in self-supervised learning~\cite{Rotation}. 
		 Image rotation can be considered as one of the image geometric transformations, which is very easy to perform. 
		 The core idea for this task is to use the neural network to estimate geometric transformation (i.e., the angle of rotation in this case). 
		A common practice is to define the set of geometric transformation $\mathcal{R}$ as all the image rotations of 90 degrees (e.g., 2d image rotations by 0, 9, 180 and 270 degrees). Here, we use the size of transformation set $\mathcal{R}$ to control the task difficulty. As shown in Table~\ref{tab:rotation}, we defined three set of geometric transformations as all image rotations of 180/90/45 degree, with 2/4/8 operations within each set respectively. For example, $\mathcal{R}_1=\{0^{\circ}, 180^{\circ}\}$, $\mathcal{R}_2=\mathcal{R}_1 \cup \{90^{\circ}, 270^{\circ}\}$ and $\mathcal{R}_3=\mathcal{R}_2 \cup \{45^{\circ}, 135^{\circ}, 225^{\circ}, 315^{\circ}\} $. Notice that, $\mathcal{R}_1 \subseteq \mathcal{R}_2\subseteq \mathcal{R}_3$, which indicates an increase in task complexity. 
	
		 \begin{table}
			\begin{center}
				\begin{tabular}{c|c|c|c}
					\hline
					Task &  Set$^\dagger$ & Cardinality& Angle Base  \\
					\hline
					$\mathcal{G}_1$ & $\mathcal{R}_1$ & 2   & $180^{\circ}$  \\
					$\mathcal{G}_2$ & $\mathcal{R}_2$ & 4   & $90^{\circ}$   \\
					$\mathcal{G}_3$ & $\mathcal{R}_3$ & 8  &$45^{\circ}$     \\
					\hline
				\end{tabular}
			\end{center}
			\caption{Multi-level task design for image rotation. Set$^\dagger$ denotes the rotation transformation set.}
			\label{tab:rotation}
		\end{table}
		
		\begin{table}
		\begin{center}
			\begin{tabular}{c|c|c}
				\hline
				Task & Set$^\dagger$ &  scheme   \\
				\hline
				$\mathcal{G}_1$ & $\mathcal{T}_1$ &  random crop   \\
				$\mathcal{G}_2$ & $\mathcal{T}_2$ & crop+color distortion      \\
				$\mathcal{G}_3$ & $\mathcal{T}_3$ & crop+color distortion+filtering  \\
				\hline
			\end{tabular}
		\end{center}
		\caption{Multi-level task design for contrastive learning. Set$^\dagger$  denotes the transformation set.}
		\label{tab:contrastive}
		\end{table}
		
	\subsubsection{Multi-level Contrastive Learning}\label{sec:task_Contrastive}
		Contrastive learning has recently become a dominant approach in the area of self-supervised learning~\cite{SimCLR, Moco, MocoV2}. Typically, contrastive learning approaches learn representations by contrasting positive pairs against negative pairs. For example, SimCLR~\cite{SimCLR} makes use of multiple data augmentation operations to generate positive data pairs. Then a based encoder network is trained to maximize the similarity of positive data pairs meanwhile minimize the similarity of negative data pairs using a contrastive loss. We design our multi-level contrastive tasks based on the setting of SimCLR~\cite{SimCLR}. We control the task difficulty by manipulating the augmentation set $\mathcal{T}$.
		For the low-level task (e.g., $\mathcal{G}_1$), we use a simple augmentation scheme and we increase the complexity of the augmentation scheme for high-level tasks  (e.g., $\mathcal{G}_3$).
		 Three kinds of augmentation are adopted: i) geometric transformation of data, such as cropping and resizing; ii) appearance transformation, such as color distortion and iii) other transformation, such as Gaussian blur and Sobel Filtering.
		 As shown in Table~\ref{tab:contrastive}, we set $\mathcal{T}_1=\{\text{Random Crop}\}$ as the first augmentation set for task $\mathcal{G}_1$ and we add color distortion into the augmentation set $,_2$ for task $\mathcal{G}_2$. Other transformation operations are included in the augmentation set $\mathcal{T}_3$ for task $\mathcal{G}_3$. Notice that $\mathcal{T}_1 \subseteq \mathcal{T}_2 \subseteq \mathcal{T}_3$.

\subsection{Stage-wise Network Training}~\label{sec:StageWiseTraining}
	As explained in section~\ref{sec:PSL}, three learning tasks $\{\mathcal{G}_1,\mathcal{G}_2,\mathcal{G}_3\}$ are obtained, corresponding to three losses $\{\mathcal{L}_1,\mathcal{L}_2,\mathcal{L}_3\}$ and stages $S_1, S_2, S_3$. Unlike traditional end-to-end learning, we adopt a local learning strategy. 
	To be specific, the gradient of each learning stage does not flow back to other stages. For example, gradients generated by $\mathcal{L}_2$ only influence layers within $S_2$ (i.e.,$B_2$, $B_3$ and $B_4$) during the second learning stage. This allow the corresponding layers focus on the current learning target, which turns out to be good for overall training.
	Our learning algorithm is summarized in Algorithm~\ref{alg:train}. After we finish multi-level task design and stage-wise network partition, we start progressive stage-wise learning. There are three learning stages in total. During the $i$-th stage, we do data pre-processing first, which is determined by the specific learning task $\mathcal{G}_i$. Then we do forward propagation with $h_k=f(x_k|S_i)$ representing the output feature of stage $S_i$. Then, $h_k$ is sent to a decoder $g_i$ for further processing before applying the stage loss $\mathcal{L}_i$. Only layers within $S_i$ and the decoder $g_i$ are updated.  After the training, all decoders will be removed and no extra computation cost is introduced in $f$.  
	
	Notice that there are overlappings between each stage. Another stage partition approach is to cut the encoder into several non-overlapping parts and train the whole network in a greedy layer-wise manner like GIM~\cite{GIM}. 
	In the case of GIM, upper layers/stages cannot receive gradient feedback from lower layers/stages.
	However, as the difficulty of the multi-level task increases, the quality of the intermediate representation of lower stages has a large influence on the final performance of the upper stages. Therefore, it is necessary for stages to have overlapping layers, which play a role in connecting and communicating between stages. 
%	\begin{algorithm}
%		\caption{An algorithm}
%		\begin{algorithmic}[1]
%			\ForEach{$a \in A$}%
%			\State command \algorithmiccomment{This is a comment}
%			\State another command \algorithmiccomment{This is another comment}
%			\EndForEach
%		\end{algorithmic}
%	\end{algorithm}

	\begin{algorithm}[t] 
		\caption{Progressive Stage-wise Learning Algorithm.} 
		\label{alg:train} 
		\begin{algorithmic}[1] 
			\Require 
			Learning Target $\mathcal{G}$, Learning Loss $\mathcal{L}$, backbone Network $f$, projection head $g_1$, $g_2$ and $g_3$, batch size $N$
			\State $\mathcal{G}\rightarrow\{\mathcal{G}_1,\mathcal{G}_2,\mathcal{G}_3\}$
			\algorithmiccomment{\textcolor[rgb]{0.5,0.5,0.5}{Multi-level Task Design}}
			\State $\mathcal{L}\rightarrow\{\mathcal{L}_1,\mathcal{L}_2,\mathcal{L}_3\}$
			\State $f\rightarrow\{S_1,S_2,S_3\}$
			\algorithmiccomment{\textcolor[rgb]{0.5,0.5,0.5}{Stage-wise Network Partition}}
			\For {$i\in\{1,2,3\}$}
				\algorithmiccomment{\textcolor[rgb]{0.5,0.5,0.5}{Stage-wise Learning}}
				\For{sampled minibatch $\{x_k\}_{k=1}^N$}
					\State Data pre-processing for task $\mathcal{G}_i$
					\State $h_k=f(x_k|S_i)$
					\algorithmiccomment{\textcolor[rgb]{0.5,0.5,0.5}{Forward propagation}}
					\State $z_k=g_i(h_k)$
					\State Computer gradient with respect to $\mathcal{L}_i(x_k, z_k)$
					\State Update layers within $S_i$ 
					\State Update $g_i$
				\EndFor
			\EndFor
			\State \textbf{Return} $f$, and discard $g_1$, $g_2$ and $g_3$
%			\Ensure 
%			Searched Architecture $g$
%			 \State $\mathcal{G} \leftarrow \emptyset$ 
%			\State Split $\mathcal{X}_\mathrm{train}$ into ($\mathcal{X}_{w}$, $\mathcal{X}_{\alpha}$) according to $\eta$
%			\State $B_w \leftarrow \lfloor |\mathcal{X}_{w}| / B \rfloor$
%			\State $B_{\alpha} \leftarrow \lfloor |\mathcal{X}_{\alpha}| / B \rfloor$
%			\For {$\mathrm{epoch}= 0 : N_{\max}$}
%			\For {$\mathrm{iter}= 0 : B_w$}
%			\State \small{$G_\mathrm{sub} \leftarrow \mathrm{SubSuperNetSampler}(G, \alpha, \mathrm{epoch})$}
%			\State Train $w$ of $G_\mathrm{sub}$ on $\mathcal{X}_{w}$ for one batch.
%			\If {\small{$\mathrm{iter}$ \% $\lfloor B_w / B_{\alpha} \rfloor == 0\ \&\ \mathrm{epoch} > N_w$}}
%			\State \small{$G_{sub} \leftarrow \mathrm{SubSuperNetSampler}(G, \alpha, \mathrm{epoch})$}
%			\State Train $\alpha$ of $G_\mathrm{sub}$ on $\mathcal{X}_{\alpha}$ for one batch.
%			\EndIf
%			\EndFor
%			% \If {epoch $> N_w$}
%			%       \State $g \leftarrow ArchitectureSampling(G, \alpha)$
%			%       \State push $g$ to $\mathcal{G}$
%			% \EndIf
%			\EndFor
%			\State $g \leftarrow \mathrm{ArchitectureSampler}(G, \alpha)$
%			\State Return $g$ 
		\end{algorithmic} 
	\end{algorithm}

\section{Experiment}
In this section, we conduct experiments to validate the effectiveness of the proposed Progressive Stage-wise Learning (PSL) framework. Firstly, we apply PSL on several different kinds of unsupervised/self-supervised learning tasks to evaluate the quality of the learned representation on ImageNet~\cite{ImageNet}. 
Then, we report experiment results of downstream tasks, i.e., semi-supervised and transfer learning. 

\subsection{Implementation Details}
	Generally, we conduct contrast experiments on three different unsupervised/self-supervised learning methods. The basic implementation details for each task as follows:
	
        \vspace{-0.3cm}
		\paragraph{Jigsaw puzzle.}
		We perform multi-level task design  $\mathcal{G}\rightarrow\{\mathcal{G}_1,\mathcal{G}_2,\mathcal{G}_3\}$  following Sec~\ref{sec:task_Jigsaw}. Correspondingly, we use three different cardinality $\{500, 1000, 2000\}$ for the permutation set $\{\mathcal{C}_1, \mathcal{C}_2, \mathcal{C}_3\}$, with $\mathcal{C}_1 \subseteq\mathcal{C}_2\subseteq\mathcal{C}_3$, representing an increase in task complexity.
		We use ResNet-50~\cite{ResNet} as our backbone network and do PSL training on 8-gpus. For each input image, we resize the shorter side to resolution 256, randomly crop a 225x225 image and apply horizontal flip with 50\% probability. For the training, we use mini-batch size of 256, initial learning rate of 0.01 with the learning rate dropped by a factor of 10. Following~\cite{Scaling}, we use momentum of 0.9, weight decay 1e-4 with no decay for the bias parameters. 
		For each training stage of PSL, we train for 60 epochs and use the learning rate schedule of 20/20/10/10. 
		
		\vspace{-0.3cm}
		\paragraph{Image rotation.}
		We perform multi-level task design $\mathcal{G}\rightarrow\{\mathcal{G}_1,\mathcal{G}_2,\mathcal{G}_3\}$  following Sec~\ref{sec:task_Rotation}. Correspondingly, we set the rotation transformation set $\mathcal{R}_1=$$\{0^{\circ},  180^{\circ}\}$, $\mathcal{R}_2=$$\{(90\times i)^{\circ}|0\leq i\leq 3\}$ and  $\mathcal{R}_3=$$\{(45\times i)^{\circ}|0\leq i\leq 7\}$. 
		Notice that $\mathcal{R}_1 \subseteq \mathcal{R}_2\subseteq \mathcal{R}_3$, representing an increase in task complexity.
		We use RevNet50~\cite{RevNet} as our backbone network and do PSL training on 8-gpus\footnote{Compared with ResNet~\cite{ResNet}, RevNet~\cite{RevNet} is more suitable in image rotation tasks~\cite{Revisiting}. We don't use RevNet50w4$\times$, which is reported with better performance, because scaling up model complexity is not discussed in this paper.}. For each input image, we resize the shorter side to 256 and do the rotation transformation according to the transformation set. We perform a center crop on the rotated image remaining the resolution and then 
		randomly crop 224x224 image. For the training, we use mini-batch size of 256, initial learning rate of 0.01 with the learning rate dropped by a factor of 10. 
		For each training stage of PSL, we train for 60 epochs and use the learning rate schedule of 20/20/10/10. 
		
		\vspace{-0.3cm}
		\paragraph{Contrastive learning.}
		We perform multi-level task design $\mathcal{G}\rightarrow\{\mathcal{G}_1,\mathcal{G}_2,\mathcal{G}_3\}$ following Sec~\ref{sec:task_Contrastive}. Correspondingly, we set the transformation set as: $\mathcal{T}_1=$\{Random crop\}, $\mathcal{T}_2=\mathcal{T}_1\cup $ \{Color Distortion\} and  $\mathcal{T}_3=\mathcal{T}_2\cup $ \{Gaussian Blur, Sobel Filtering\}. Notice that $\mathcal{T}_1 \subseteq \mathcal{T}_2\subseteq \mathcal{T}_3$, representing an increase in task complexity.
		We use ResNet-50 as our backbone network and a 2-layer MLP projection head to project the representation to a 128-dimensional space. Limited by computing resources, we use mini-batch size of 256 and train for 200 epochs (60/70/70 epochs for each learning stage) on 8-gpus. Notice that the max performance is not obtained in 200 epochs and 256 batch-size\footnote{1000 epochs with batch-size 4096 is reported with the best performance in SimCLR~\cite{SimCLR}.}, reasonable results and fair comparison can still be achieved. Based on SimCLR~\cite{SimCLR}, We use NT-Xent loss, learned in LARS optimizer with learnig rate of 0.3 with a cosine decay schedule without restart. Similarly, we apply PSL on the data-augmentation part of MoCov2~\cite{MocoV2} and get an improvement of 0.6\%.

\begin{table}[t]
	\begin{center}
		\begin{tabular}{ccccc}
			\toprule[1.5pt]
			Method  & Arch &  \# Param(M)  & Top 1 \\
			\hline
			Colorization~\cite{Colorization2016A} &  R50 & 24  & 39.6 \\
			BigBiGAN~\cite{BigBiGAN}  	&  R50 & 24  & 56.6 \\
			LA~\cite{LA2019} &  R50 & 24  & 58.8\\
			NPID++~\cite{misra2020self}&  R50 & 24  & 59.0 \\
			MoCo~\cite{Moco}&  R50 & 24  & 60.6\\
			PIRL~\cite{misra2020self} &  R50 & 24  & 63.6\\
			CPC v2~\cite{CPCv2} &  R50 & 24  & 63.8\\
			PCL~\cite{PCL}      &  R50 & 24  & 65.9\\
			SwAV~\cite{SwAV} &  R50 & 24  & 75.3\\
			\hline
			Jigsaw~\cite{Jigsaw} &  R50& 24  & 45.7\\
			\textbf{Jigsaw $+$PSL}&  R50 & 24  & \textbf{50.9} \\
			\hline
			Rotation~\cite{Rotation} &  Rv50w4$\times$ & 86  & 47.3\\
			Rotation*&  Rv50& 24  & 48.6\\
			\textbf{Rotation$+$PSL}  &  Rv50 & 24 & \textbf{53.3}\\
			\hline
			SimCLR~\cite{SimCLR}  &  R50 & 24  & 61.9\\
			\textbf{SimCLR$+$PSL} &  R50 & 24  & \textbf{64.3}\\
			\hline
			MoCov2~\cite{MocoV2} &  R50 & 24  & 67.5\\
			\textbf{MoCov2+PSL} & R50 & 24 & \textbf{68.1}\\
			\bottomrule[1.5pt]
		\end{tabular}
	\end{center}
	\caption{ImageNet accuracy of linear classifiers trained on self-supervised learned representations. All are reported as unsupervised pre-training on ImageNet, followed by supervised linear classification and evaluated on the ImageNet validation set. Note that Rotation~\cite{Rotation} uses $\mathcal{R}_2$ as the transformation set while Rotation* uses $\mathcal{R}_3$ as the transformation set. SimCLR results are obtained by 200 training epochs with batchsize 256.}
	\label{tab:linear}
\end{table}

\subsection{Linear Classification}

	\begin{table}[t]
		\begin{center}
			\begin{tabular}{c|c|ccccc}
				\toprule[1.5pt]
				\textbf{Method}  & \textbf{Arch} &  $B_1$ & $B_2$ & $B_3$ & $B_4$ & $B_5$\\
				\midrule[0.8pt]
				Supervised & R50 & 11.6 & 33.3 & 48.7 & 67.9 & 75.5\\
				\midrule[0.8pt]
				Jigsaw	       & R50 & 12.4 & 28.0 & 39.9 & 45.7 & 34.2\\
				SL				   &R50 & 10.8& 27.2& 39.6& 45.5 & 34.7\\
				\textbf{PSL}  & R50 & 10.9 & 27.0 & 43.2 & \textbf{50.9} & 38.5\\
				\midrule[0.8pt]
				$\text{PSL}_f$	&R50 & 10.8 & 27.3 & 43.1& 51.0 & 38.2\\
				\bottomrule[1.5pt]
			\end{tabular}
		\end{center}
		\caption{ Jigsaw Puzzle task of ResNet-50 top-1 centrer-crop accuracy for linear classification on the ImageNet dataset. Here $B_1\sim B_5$ represent the blocks defined in Sec~\ref{sec:PSL}. The supervised results are presented for reference. Jigsaw is end-to-end trained with the task $\mathcal{G}_3$. SL is short for stage-wise learning. In SL, we do stage-wise training with the jiasaw puzzle task $\mathcal{G}_3$ as the learning target of each stage. $\text{PSL}_f$ is a full-gradient version of PSL. More discussion can be found in ablation in Sec~\ref{sec:analysis}.}
		\label{tab:linear_jigsaw}
	\end{table}

	\begin{table}[t]
		\begin{center}
			\begin{tabular}{c|c|ccccc}
				\toprule[1.5pt]
				\textbf{Method}  & \textbf{Arch} &  $B_1$ & $B_2$ & $B_3$ & $B_4$ & $B_5$\\
				\midrule[0.8pt]
				Supervised & Rv50 & 11.7 & 32.6 &  47.8& 66.6 & 74.3\\
				\midrule[0.8pt]
				Rotation	       & Rv50 & 10.9 & 30.1 & 40.2 & 48.6 & 46.5 \\
				SL  & Rv50 & 11.1 &  29.1& 41.5 &  50.7 & 47.9\\
				\textbf{PSL}  & Rv50 & 11.3 &  30.8& 42.9 & \textbf{53.3} & 49.5\\
				\midrule[0.8pt]
				$\text{PSL}_f$	&R50 & 10.5 & 31.1 & 42.1& 52.9 & 49.7\\
				\bottomrule[1.5pt]
			\end{tabular}
		\end{center}
		\caption{Image rotation task of RevNet-50 top-1 centrer-crop accuracy for linear classification on the ImageNet dataset. Here $B_1\sim B_5$ represent the block defined in Sec~\ref{sec:PSL}. The supervised results are presented for reference. Rotation is end-to-end trained with the task $\mathcal{G}_3$. SL is short for stage-wise learning.  In SL, we do stage-wise training with the image rotation task $\mathcal{G}_3$ as the learning target of each stage. $\text{PSL}_f$ is a full-gradient version of PSL. More discussion can be found in ablation in Sec~\ref{sec:analysis}.}
		\label{tab:linear_rotation}
	\end{table}

In this subsection, we verify our method by linear classification on ImageNet~\cite{ImageNet}.
%\textbf{ImageNet}~\cite{ImageNet} consists of $\sim$1.28 million natural images and 50K validation images, including 1000 categories. This dataset is well-balanced in its class distribution.
Following a common protocol, We conduct contrast experiments on three different unsupervised/self-supervised learning methods. Firstly, we perform unsupervised pre-training on ImageNet. Then, we train a supervised linear classifier (a fully-connect layer followed by softmax). Table~\ref{tab:linear} summaries the single crop top-1 classification accuracy on the ImageNet validation set, comparing our results with previous approaches~\cite{Colorization2016A,BigBiGAN,LA2019,misra2020self,Moco,CPCv2,PCL,MocoV2,SwAV} as well as three baseline method~\cite{Jigsaw, Rotation, SimCLR}. 
Compared with the vanilla jigsaw puzzle task, PSL training improves the results by 5.2\% (45.7\%$\rightarrow$ 50.9\%). Compared with the vanilla image rotation task, PSL training improves the results by 4.7\%  (48.6\%$\rightarrow$ 53.3\%). As for the SimCLR contrastive learning, PSL training improves the results by 2.4\%  (61.9\% $\rightarrow$ 64.3\%), under the setting of 200 training epoch and batch-size 256. These results indicate the PSL training can effectively improve the unsupervised learning quality. Specifically, we extract image features from five different layers (i.e., the output of $B_1$, $B_2$, $B_3$, $B_4$ and $B_5$) after unsupervised pre-training and train linear classifiers on these fixed representations. 
For the jigsaw puzzle task, detailed results are shown in Table~\ref{tab:linear_jigsaw}. Results of the image rotation task 
are presented in Table~\ref{tab:linear_rotation}.

\subsection{Semi-supervised Learning}

	\begin{table}[t]
		\begin{center}
			\begin{tabular}{ccccc}
				\toprule[1.5pt]
				Task	&	 1\% labels & 10\% labels\\
				\hline
				Supervised &  48.4  & 80.4 \\
				\hline
				Jigsaw~\cite{Jigsaw} & 45.4  & 79.6  \\
				Jigsaw+PSL	 &	\textbf{48.7}  &\textbf{ 83.5} \\
				\hline
				Rotation~\cite{Rotation} & 52.1  &  82.5  \\
				Rotation+PSL  & \textbf{ 54.8}  & \textbf{ 83.7}  \\
				\bottomrule[1.5pt]
			\end{tabular}
		\end{center}
		\caption{\textbf{Semi-supervised learning on ImageNet.} We use ResNet-50 as our backbone networks and report single-crop top-5 accuracy on the ImageNet validation set. All models are self-supervised trained on ImageNet and finetuned on 1\% and 10\% of the ImageNet training data, following~\cite{wu2018unsupervised, misra2020self}. The supervised results are presented for reference.}
		\label{tab:semi}
	\end{table}

Following the experiment setup of~\cite{wu2018unsupervised, misra2020self},  we perform semi-supervised image classification on ImageNet to evaluate the effectiveness of the pre-trained network in data-efficient settings. We do a class-balanced data selection to obtain 1\% and 10\% of the ImageNet training data and finetuned the whole pre-trained network. Table~\ref{tab:semi} reports the top-5 accuracy of the resulting models on the ImageNet validation set.
By contrast experiments on two pretext tasks, the proposed PSL training method can improve the top-5 accuracy by a large margin. In the jigsaw puzzle task, PSL brings an improvement of 3.3\% and 3.9\% for 1\% and 10\% labeled data respectively. For the image rotation task, PSL improves the performance by 2.7\% and 1.2\% respectively. 

\subsection{Transfer Learning}

To further investigate the generalization ability of our method, we conduct transfer learning experiments including object detection on PASCAL VOC~\cite{PascalVOC} and image classification results on three datasets. In this subsection, we use ResNet-50 as our backbone network both jigsaw puzzle and image roration tasks. 
Results are reported in Table~\ref{tab:transfer} and Table~\ref{tab:transfer2}. 

\subsubsection{Object Detection}

	\begin{table}[t]
		\begin{center}
			\begin{tabular}{ccccc}
				\toprule[1.5pt]
				\multirow{2}{*}{Task}		&	\multicolumn{4}{c}{Object Detection} \\
				\cline{2-5}
				& AP$_{\text{all}}$  	&	AP$_{\text{50}}$  & AP$_{\text{75}}$  & $\Delta\text{AP}_{\text{75}}$\\
				\hline
				Supervised & 52.6&81.1&57.4& 0.0\\
				\hline
				Jigsaw~\cite{Jigsaw} & 48.9 & 75.1 &  52.9 & -4.5\\
				Jigsaw+PSL		 &	\textbf{51.1}&\textbf{ 78.9}&  \textbf{55.3} &  \textbf{-2.1}\\
				\hline
				Rotation~\cite{Rotation}  &46.3 & 72.5 &  49.3 & -8.1\\
				Rotation+PSL  	& \textbf{47.6} & \textbf{74.6}&\textbf{51.1}&\textbf{ -6.3}\\
				\bottomrule[1.5pt]
			\end{tabular}
		\end{center}
		\caption{Object detection results of transfer learning on PASCAL VOC dataset. We report AP on the test set after finetuning Faster R-CNN modelswith a ResNet-50 backbone, that are unsupervised pre-trained  on ImageNet. The supervised results are presented for reference.}
		\label{tab:transfer}
	\end{table}

 Following previous works~\cite{Scaling, misra2020self}, we perform object detection experiments on the PASCAL VOC dataset~\cite{PascalVOC} using VOC07+12 training split. We use Faster RCNN~\cite{FasterRCNN} object detector and ResNet-50 C4 ~\cite{ResNet} backbone. We follow the same training schedule as~\cite{Scaling, misra2020self} to finetune all models on VOC with BatchNorm parameters fixed during finetuning. We report our performance of \{AP$_{\text{50}}$, AP$_{\text{75}}$, $\Delta\text{AP}_{\text{75}}$\} in Table~\ref{tab:transfer}.
 Compared with the vanilla pretext task, our PSL training scheme can substantially improve the three indicators, representing an enhancement of the learned feature representation. In the jigsaw puzzle pretext task, we get an improvement of \{2.2, 3.8, 2.4\} respectively, while in the image rotation pretext task, PSL brings an improvement of \{1.3, 2.1, 1.8\}. 
\subsubsection{Image Classification on other datasets}

	\begin{table}[t]
		\begin{center}
			\begin{tabular}{ccccc}
				\toprule[1.5pt]
				\multirow{2}{*}{Task}		&	\multicolumn{3}{c}{Transfer Dataset} \\
				\cline{2-4}
				& PASCAL  & Places & iNat18 \\
				\hline
				Supervised & 87.5 & 51.5  & 45.4\\
				\hline
				Jigsaw~\cite{Jigsaw}  & 64.5 &  41.2 & 21.3 \\
				Jigsaw+PSL	 &	\textbf{67.2}& \textbf{44.7}  & \textbf{24.1} \\
				\hline
				Rotation~\cite{Rotation}   & 63.5 &  41.9 & 23.0 \\
				Rotation+PSL  & \textbf{65.9} & \textbf{43.0} & \textbf{25.9} \\
				\bottomrule[1.5pt]
			\end{tabular}
		\end{center}
		\caption{Image classification results of transfer learning on PASCAL~\cite{PascalVOC}, Places~\cite{Places} and iNat18~\cite{iNat}. We use linear classifiers on image representations obtained by self-supervised learners that are pre-trained on ImageNet. We report mAP on the PASCAL dataset and top-1 accuracy on Places and iNat18. We compare results on the jigsaw puzzle and image rotation pretext tasks with the proposed PSL. The supervised results are presented for reference.}
		\label{tab:transfer2}
	\end{table}

 Next, we conduct transfer learning experiments on the image classification task. We use models pre-trained on ImageNet and assess the quality of learned features by training linear classifiers on fixed image representations. 
  Following the setting of~\cite{Scaling}, we evaluate feature representations extracted from five intermediate blocks of the pre-trained network, and report the best classification results in Table~\ref{tab:transfer2}. We report transfer learning performance on PASCAL~\cite{PascalVOC}, Places~\cite{Places} and iNat18~\cite{iNat}. 
%   \textbf{Places}~\cite{Places} dataset contains 2.4M training images and 20.5K validation images, including 205 scene categories.
%   \textbf{PASCAL}~\cite{PascalVOC} classification involves 20 binary classificaiton tasks. We report AP for this dataset.
%   \textbf{iNat}~\cite{iNat} dataset contains 8000 species, with a combined training and validation set of 450K images.
  In PASCAL, our PSL training improves the jigsaw puzzle task by 2.7\%, and image rotation task by 2.4\%. In Places, the PSL training improves the performance by 3.5\% and 1.1\%  while in iNat18, the improvement is 2.8\% and 2.9\% for these two tasks respectively.

\subsection{Analysis}~\label{sec:analysis}
	\vspace{-0.8cm}
	\paragraph{Ablation: Progressive Mechanism.} Here we discuss the effectiveness of the progressive mechanism. We design comparison experiments in ImageNet linear classification to compare stage-wise learning w/o progressive learning, namely \textbf{PSL} \textit{vs.} \textbf{SL}. For SL, we use the same learning task ($\mathcal{G}_3$) in each learning stage, which means the network is trained to learn the hardest task for each learning stage. For PSL, the task complexity increases for learning stages $S_1$, $S_2$ and $S_3$. We present PSL \textit{vs. } SL in Table~\ref{tab:linear_jigsaw} and Table~\ref{tab:linear_rotation}. For both tasks, PSL leads to better results than SL without the progressive mechanism.

    \vspace{-0.2cm}
	\paragraph{Ablation: Gradient Association.} As discussed in Section~\ref{sec:StageWiseTraining}, we adopt a stage-wise training scheme with limited a gradient association in each learning stage. Specifically, the gradient of $\mathcal{L}_2$ will not flow back to $B_1$ and the gradient of $\mathcal{L}_3$ will not influence $B_1$ and $B_2$.
	We implement the full gradient version of PSL where in each learning stage, all layers are updated without gradient restriction. We show the linear classification results on ImageNet dataset as $\text{PSL}_f$ in Table~\ref{tab:linear_jigsaw} and Table~\ref{tab:rotation}.  From the results, we conclude that full gradient training does not necessarily bring an improvement in the unsupervised training performance. 
	Therefore, we set gradient restriction in each learning stage, which will also reduce the computation cost during training. 
	
	\vspace{-0.2cm}
	\paragraph{Local Training.} We do not adopt a greedy block-wise learning scheme like~\cite{GIM}. Instead, we enhance stage-wise the connection by enabling graident flow between stages. A similar approach is adopted in LoCo~\cite{Loco}, where gradient connections are enabled in adjacent blocks. In PSL, block connections are further enhanced (e.g., learning  stage $S_3$ have impact on $B_3$, which is also included in learning stage $S_1$). By comparing $\text{PSL}$ and $\text{PSL}_f$ in Table~\ref{tab:linear_jigsaw} and \ref{tab:linear_rotation}, we show that this design helps enhance the learning quality of various unsupervised tasks.
	
\section{Conclusion}
	In this work, we present a Progressive Stage-wise Learning (\textbf{PSL}) framework for unsupervised/self-supervised learning. Through multi-level task design and progressive stage-wise training, PSL improves many mainstream unsupervised methods. We provide experiments in three different tasks (i.e., jigsaw puzzle, image rotation and contrastive learning) in this paper and validate the effectiveness of PSL. Our future work involves exploring PSL on other tasks and searching for the optimal architecture for PSL framework. We hope PSL can be a universal unsupervised training approach in enhancing the learned feature representation. 

\vspace{-0.2cm}
\paragraph{Acknowledgements.} 
    This work was supported by National Science Foundation of China (U20B2072, 61976137), NSFC(U19B2035), Shanghai Municipal Science and Technology Major Project (2021SHZDZX0102), ONR N00014-20-1-2206 and China Scholarship Council (NO.201906230178).

\newpage
{\small
\bibliographystyle{ieee_fullname}
\bibliography{egbib}

\begin{thebibliography}{10}\itemsep=-1pt

\bibitem{asano2019self}
Yuki~Markus Asano, Christian Rupprecht, and Andrea Vedaldi.
\newblock Self-labelling via simultaneous clustering and representation
  learning.
\newblock {\em arXiv preprint arXiv:1911.05371}, 2019.

\bibitem{AMDIM}
Philip Bachman, R~Devon Hjelm, and William Buchwalter.
\newblock Learning representations by maximizing mutual information across
  views.
\newblock In {\em Adv. Neural Inform. Process. Syst.}, pages 15535--15545,
  2019.

\bibitem{belilovsky2019greedy}
Eugene Belilovsky, Michael Eickenberg, and Edouard Oyallon.
\newblock Greedy layerwise learning can scale to imagenet.
\newblock In {\em International Conference on Machine Learning}, pages
  583--593. PMLR, 2019.

\bibitem{bengio2006greedy}
Yoshua Bengio, Pascal Lamblin, Dan Popovici, and Hugo Larochelle.
\newblock Greedy layer-wise training of deep networks.
\newblock {\em Adv. Neural Inform. Process. Syst.}, 19:153--160, 2006.

\bibitem{Curriculum}
Yoshua Bengio, J{\'e}r{\^o}me Louradour, Ronan Collobert, and Jason Weston.
\newblock Curriculum learning.
\newblock In {\em International conference on machine learning}, pages 41--48,
  2009.

\bibitem{Jigsaw2019}
Fabio~M Carlucci, Antonio D'Innocente, Silvia Bucci, Barbara Caputo, and
  Tatiana Tommasi.
\newblock Domain generalization by solving jigsaw puzzles.
\newblock In {\em IEEE Conf. Comput. Vis. Pattern Recog.}, pages 2229--2238,
  2019.

\bibitem{caron2018deep}
Mathilde Caron, Piotr Bojanowski, Armand Joulin, and Matthijs Douze.
\newblock Deep clustering for unsupervised learning of visual features.
\newblock In {\em Eur. Conf. Comput. Vis.}, pages 132--149, 2018.

\bibitem{caron2019unsupervised}
Mathilde Caron, Piotr Bojanowski, Julien Mairal, and Armand Joulin.
\newblock Unsupervised pre-training of image features on non-curated data.
\newblock In {\em Int. Conf. Comput. Vis.}, pages 2959--2968, 2019.

\bibitem{SwAV}
Mathilde Caron, Ishan Misra, Julien Mairal, Priya Goyal, Piotr Bojanowski, and
  Armand Joulin.
\newblock Unsupervised learning of visual features by contrasting cluster
  assignments.
\newblock {\em Adv. Neural Inform. Process. Syst.}, 33, 2020.

\bibitem{SimCLR}
Ting Chen, Simon Kornblith, Mohammad Norouzi, and Geoffrey~E. Hinton.
\newblock A simple framework for contrastive learning of visual
  representations.
\newblock {\em CoRR}, abs/2002.05709, 2020.

\bibitem{MocoV2}
Xinlei Chen, Haoqi Fan, Ross Girshick, and Kaiming He.
\newblock Improved baselines with momentum contrastive learning.
\newblock {\em arXiv preprint arXiv:2003.04297}, 2020.

\bibitem{ImageNet}
Jia Deng, Wei Dong, Richard Socher, Li-Jia Li, Kai Li, and Li Fei-Fei.
\newblock Imagenet: A large-scale hierarchical image database.
\newblock In {\em IEEE Conf. Comput. Vis. Pattern Recog.}, pages 248--255.
  Ieee, 2009.

\bibitem{Colorization2015}
Aditya Deshpande, Jason Rock, and David Forsyth.
\newblock Learning large-scale automatic image colorization.
\newblock In {\em Int. Conf. Comput. Vis.}, pages 567--575, 2015.

\bibitem{ContextPrediction}
Carl Doersch, Abhinav Gupta, and Alexei~A Efros.
\newblock Unsupervised visual representation learning by context prediction.
\newblock In {\em Int. Conf. Comput. Vis.}, pages 1422--1430, 2015.

\bibitem{BigBiGAN}
Jeff Donahue and Karen Simonyan.
\newblock Large scale adversarial representation learning.
\newblock In {\em Adv. Neural Inform. Process. Syst.}, pages 10542--10552,
  2019.

\bibitem{Contrastive2014}
Alexey Dosovitskiy, Jost~Tobias Springenberg, Martin Riedmiller, and Thomas
  Brox.
\newblock Discriminative unsupervised feature learning with convolutional
  neural networks.
\newblock In {\em Adv. Neural Inform. Process. Syst.}, pages 766--774, 2014.

\bibitem{PascalVOC}
Mark Everingham, SM~Ali Eslami, Luc Van~Gool, Christopher~KI Williams, John
  Winn, and Andrew Zisserman.
\newblock The pascal visual object classes challenge: A retrospective.
\newblock {\em Int. J. Comput. Vis.}, 111(1):98--136, 2015.

\bibitem{selfpaced4}
Hongchang Gao and Heng Huang.
\newblock Self-paced network embedding.
\newblock In {\em Proceedings of the 24th ACM SIGKDD International Conference
  on Knowledge Discovery \& Data Mining}, pages 1406--1415, 2018.

\bibitem{selfpaced2}
Yixiao Ge, Dapeng Chen, Feng Zhu, Rui Zhao, and Hongsheng Li.
\newblock Self-paced contrastive learning with hybrid memory for domain
  adaptive object re-id.
\newblock {\em arXiv preprint arXiv:2006.02713}, 2020.

\bibitem{gidaris2020learning}
Spyros Gidaris, Andrei Bursuc, Nikos Komodakis, Patrick P{\'e}rez, and Matthieu
  Cord.
\newblock Learning representations by predicting bags of visual words.
\newblock In {\em IEEE Conf. Comput. Vis. Pattern Recog.}, pages 6928--6938,
  2020.

\bibitem{Rotation}
Spyros Gidaris, Praveer Singh, and Nikos Komodakis.
\newblock Unsupervised representation learning by predicting image rotations.
\newblock In {\em Int. Conf. Learn. Represent.}, 2018.

\bibitem{RevNet}
Aidan~N Gomez, Mengye Ren, Raquel Urtasun, and Roger~B Grosse.
\newblock The reversible residual network: Backpropagation without storing
  activations.
\newblock In {\em Adv. Neural Inform. Process. Syst.}, pages 2214--2224, 2017.

\bibitem{Scaling}
Priya Goyal, Dhruv Mahajan, Abhinav Gupta, and Ishan Misra.
\newblock Scaling and benchmarking self-supervised visual representation
  learning.
\newblock In {\em Int. Conf. Comput. Vis.}, pages 6391--6400, 2019.

\bibitem{selfpaced1}
Xifeng Guo, Xinwang Liu, En Zhu, Xinzhong Zhu, Miaomiao Li, Xin Xu, and
  Jianping Yin.
\newblock Adaptive self-paced deep clustering with data augmentation.
\newblock {\em IEEE Transactions on Knowledge and Data Engineering},
  32(9):1680--1693, 2019.

\bibitem{Contrastive2006}
Raia Hadsell, Sumit Chopra, and Yann LeCun.
\newblock Dimensionality reduction by learning an invariant mapping.
\newblock In {\em IEEE Conf. Comput. Vis. Pattern Recog.}, volume~2, pages
  1735--1742. IEEE, 2006.

\bibitem{Moco}
Kaiming He, Haoqi Fan, Yuxin Wu, Saining Xie, and Ross Girshick.
\newblock Momentum contrast for unsupervised visual representation learning.
\newblock In {\em IEEE Conf. Comput. Vis. Pattern Recog.}, pages 9729--9738,
  2020.

\bibitem{ResNet}
Kaiming He, Xiangyu Zhang, Shaoqing Ren, and Jian Sun.
\newblock Deep residual learning for image recognition.
\newblock In {\em IEEE Conf. Comput. Vis. Pattern Recog.}, pages 770--778,
  2016.

\bibitem{CPCv2}
Olivier~J H{\'e}naff, Aravind Srinivas, Jeffrey De~Fauw, Ali Razavi, Carl
  Doersch, SM Eslami, and Aaron van~den Oord.
\newblock Data-efficient image recognition with contrastive predictive coding.
\newblock {\em arXiv preprint arXiv:1905.09272}, 2019.

\bibitem{huang2019unsupervised}
Jiabo Huang, Qi Dong, Shaogang Gong, and Xiatian Zhu.
\newblock Unsupervised deep learning by neighbourhood discovery.
\newblock {\em arXiv preprint arXiv:1904.11567}, 2019.

\bibitem{Colorization2016B}
Satoshi Iizuka, Edgar Simo-Serra, and Hiroshi Ishikawa.
\newblock Let there be color! joint end-to-end learning of global and local
  image priors for automatic image colorization with simultaneous
  classification.
\newblock {\em ACM Trans. Graph.}, 35(4):1--11, 2016.

\bibitem{Revisiting}
Alexander Kolesnikov, Xiaohua Zhai, and Lucas Beyer.
\newblock Revisiting self-supervised visual representation learning.
\newblock In {\em IEEE Conf. Comput. Vis. Pattern Recog.}, pages 1920--1929,
  2019.

\bibitem{Colorization2016C}
Gustav Larsson, Michael Maire, and Gregory Shakhnarovich.
\newblock Learning representations for automatic colorization.
\newblock In {\em Eur. Conf. Comput. Vis.}, pages 577--593. Springer, 2016.

\bibitem{Colorization2017}
Gustav Larsson, Michael Maire, and Gregory Shakhnarovich.
\newblock Colorization as a proxy task for visual understanding.
\newblock In {\em IEEE Conf. Comput. Vis. Pattern Recog.}, pages 6874--6883,
  2017.

\bibitem{PCL}
Junnan Li, Pan Zhou, Caiming Xiong, Richard Socher, and Steven~CH Hoi.
\newblock Prototypical contrastive learning of unsupervised representations.
\newblock {\em arXiv preprint arXiv:2005.04966}, 2020.

\bibitem{linsker1988self}
Ralph Linsker.
\newblock Self-organization in a perceptual network.
\newblock {\em Computer}, 21(3):105--117, 1988.

\bibitem{GIM}
Sindy Lowe, Peter O'Connor, and Bastiaan~S. Veeling.
\newblock Putting an end to end-to-end: Gradient-isolated learning of
  representations.
\newblock In {\em Adv. Neural Inform. Process. Syst.}, pages 3033--3045, 2019.

\bibitem{misra2020self}
Ishan Misra and Laurens van~der Maaten.
\newblock Self-supervised learning of pretext-invariant representations.
\newblock In {\em IEEE Conf. Comput. Vis. Pattern Recog.}, pages 6707--6717,
  2020.

\bibitem{Jigsaw}
Mehdi Noroozi and Paolo Favaro.
\newblock Unsupervised learning of visual representations by solving jigsaw
  puzzles.
\newblock In {\em Eur. Conf. Comput. Vis.}, volume 9910 of {\em Lecture Notes
  in Computer Science}, pages 69--84, 2016.

\bibitem{Count}
Mehdi Noroozi, Hamed Pirsiavash, and Paolo Favaro.
\newblock Representation learning by learning to count.
\newblock In {\em Int. Conf. Comput. Vis.}, pages 5898--5906, 2017.

\bibitem{CPC}
Aaron van~den Oord, Yazhe Li, and Oriol Vinyals.
\newblock Representation learning with contrastive predictive coding.
\newblock {\em arXiv preprint arXiv:1807.03748}, 2018.

\bibitem{FasterRCNN}
Shaoqing Ren, Kaiming He, Ross Girshick, and Jian Sun.
\newblock Faster r-cnn: Towards real-time object detection with region proposal
  networks.
\newblock In {\em Adv. Neural Inform. Process. Syst.}, pages 91--99, 2015.

\bibitem{selfpaced3}
Enver Sangineto, Moin Nabi, Dubravko Culibrk, and Nicu Sebe.
\newblock Self paced deep learning for weakly supervised object detection.
\newblock {\em IEEE transactions on pattern analysis and machine intelligence},
  41(3):712--725, 2018.

\bibitem{Inception}
Christian Szegedy, Wei Liu, Yangqing Jia, Pierre Sermanet, Scott Reed, Dragomir
  Anguelov, Dumitru Erhan, Vincent Vanhoucke, and Andrew Rabinovich.
\newblock Going deeper with convolutions.
\newblock In {\em IEEE Conf. Comput. Vis. Pattern Recog.}, pages 1--9, 2015.

\bibitem{tian2019contrastive}
Yonglong Tian, Dilip Krishnan, and Phillip Isola.
\newblock Contrastive multiview coding.
\newblock {\em arXiv preprint arXiv:1906.05849}, 2019.

\bibitem{iNat}
Grant Van~Horn, Oisin Mac~Aodha, Yang Song, Yin Cui, Chen Sun, Alex Shepard,
  Hartwig Adam, Pietro Perona, and Serge Belongie.
\newblock The inaturalist species classification and detection dataset.
\newblock In {\em IEEE Conf. Comput. Vis. Pattern Recog.}, pages 8769--8778,
  2018.

\bibitem{wu2018unsupervised}
Zhirong Wu, Yuanjun Xiong, Stella~X Yu, and Dahua Lin.
\newblock Unsupervised feature learning via non-parametric instance
  discrimination.
\newblock In {\em IEEE Conf. Comput. Vis. Pattern Recog.}, pages 3733--3742,
  2018.

\bibitem{xie2016unsupervised}
Junyuan Xie, Ross Girshick, and Ali Farhadi.
\newblock Unsupervised deep embedding for clustering analysis.
\newblock In {\em International Conference on Machine Learning}, pages
  478--487, 2016.

\bibitem{Loco}
Yuwen Xiong, Mengye Ren, and Raquel Urtasun.
\newblock Loco: Local contrastive representation learning.
\newblock {\em Adv. Neural Inform. Process. Syst.}, 33, 2020.

\bibitem{Colorization2016A}
Richard Zhang, Phillip Isola, and Alexei~A Efros.
\newblock Colorful image colorization.
\newblock In {\em Eur. Conf. Comput. Vis.}, pages 649--666. Springer, 2016.

\bibitem{Places}
Bolei Zhou, Agata Lapedriza, Jianxiong Xiao, Antonio Torralba, and Aude Oliva.
\newblock Learning deep features for scene recognition using places database.
\newblock In {\em Adv. Neural Inform. Process. Syst.}, pages 487--495, 2014.

\bibitem{LA2019}
Chengxu Zhuang, Alex~Lin Zhai, and Daniel Yamins.
\newblock Local aggregation for unsupervised learning of visual embeddings.
\newblock In {\em Int. Conf. Comput. Vis.}, pages 6002--6012, 2019.

\end{thebibliography}
}

\end{document}